\documentclass[letterpaper, 10pt, joutnal, twoside]{IEEEtran}       
\IEEEoverridecommandlockouts                              
\usepackage{graphics} 
\usepackage{epsfig} 
\usepackage{multirow}
\usepackage{color}
\usepackage{comment}

\title{Motion Switching with Sensory and Instruction Signals by designing Dynamical Systems using Deep Neural Network}

\author{Kanata Suzuki$^{1}$, Hiroki Mori$^{2}$ and Tetsuya Ogata$^{2,3}$
\thanks{$^{1}$Kanata Suzuki is with Artificial Intelligence Laboratories, Fujitsu Laboratories LTD., Kanagawa, Japan {\tt\footnotesize suzuki@idr.ias.sci.waseda.ac.jp}}
\thanks{$^{2}$Hiroki Mori and $^{2,3}$Tetsuya Ogata are with Department of Intermedia Art and Science, School of Fundamental Science and Engineering, Waseda University, Tokyo, Japan. $^{2,3}$Tetsuya Ogata are also with Artificial Intelligence Research Center, AIST, Japan {\tt\footnotesize mori@idr.ias.sci.waseda.ac.jp}; {\tt\footnotesize ogata@waseda.jp}}
}

\begin{document}


\maketitle
\begin{abstract}
To ensure that a robot is able to accomplish an extensive range of tasks, it is necessary to achieve a flexible combination of multiple behaviors. 
This is because the design of task motions suited to each situation would become increasingly difficult as the number of situations and the types of tasks performed by them increase. 
To handle the switching and combination of multiple behaviors, we propose a method to design dynamical systems based on point attractors that accept (i) "instruction signals" for instruction-driven switching. 
We incorporate the (ii) "instruction phase" to form a point attractor and divide the target task into multiple subtasks. 
By forming an instruction phase that consists of point attractors, the model embeds a subtask in the form of trajectory dynamics that can be manipulated using sensory and instruction signals. 
Our model comprises two deep neural networks: a convolutional autoencoder and a multiple time-scale recurrent neural network. 
In this study, we apply the proposed method to manipulate soft materials. 
To evaluate our model, we design a cloth-folding task that consists of four subtasks and three patterns of instruction signals, which indicate the direction of motion. 
The results depict that the robot can perform the required task by combining subtasks based on sensory and instruction signals. 
And, our model determined the relations among these signals using its internal dynamics. 
\end{abstract}


\section{INTRODUCTION}
\IEEEPARstart{T}{o}  handle soft materials, such as clothes while performing a complex task, a robot is required to not only conduct multiple subtasks properly but also to adapt to the deformation and displacement of the materials. 
These subjects are very important from the scientific and practical viewpoints. 
Soft materials have been used as objects in manipulation tasks, and a few researchers have attempted to research \cite{towel1}\cite{towel2}. 
Further, in factories and workshops, workers often perform complicated tasks that involve combining appropriate operations according to the instructions from other workers and also based on the object. 
To accomplish an extensive range of tasks, robots are also expected to perform new tasks by combining subtasks as parts of a task sequence.
\par
However, model-based methods depict some limitations in  terms of handling such  objects. Given that human workers design motions corresponding to each task, it becomes increasingly difficult to design task motions suited to each situation as the number of situations and types of tasks increase. It is especially difficult to design  flexible objects based on the conventional control theory. A significant cost is incurred with designing objects using simulators and using image processing for feature extraction  \cite{towel1}\cite{towel2}\cite{SURF}.
\par
In addition,  to switch and combine various subtasks, the method must be capable of motion branching.  In a few situations, executable subtasks cannot be determined uniquely using sensory signals alone such as the camera image. In such situations, (i) other signals to instruct the subtasks and (ii) a switching system that accepts the signals are required. The robot receives an external instruction and switches between  subtasks. However, as the task complexity increases, it becomes increasingly difficult to design all required patterns. Furthermore, creating signals in strict accordance with the circumstances degrades model versatility. Thus, the method requires a switching system that incorporates sensor- and instruction-driven switching to handle combinations of multiple types of signals.
\par
Learning-based methods are promising candidates to perform tasks that are difficult to design using model-based methods. They allow the automated acquisition of robotic motion skills. Recently, deep neural network (DNN) has been attracting considerable attention from the viewpoint of being applied to robot manipulation systems.  DNN can self-organize and extract useful  low-dimensional features from a large amount of  high-dimensional data  for diverse applications \cite{DL1}\cite{DL2}\cite{DL3}.  These features compensate for the limitations of model-based methods by autonomously extracting features of diverse environments and making generalizations. Thus, various tasks can be learned with the same model because the need to strictly design task motions and objects is eliminated. Another advantage of DNN is the ability to handle high-dimensional sensory signals without preprocessing, which enables the robot to generate and adjust tasks based on feedback derived from the images captured in real-time. 
\par
The effectiveness of the learning-based method for designing dynamical systems for the manipulation of soft materials has been confirmed as well \cite{koma}. The method can be used to generalize the object position and shape without any design effort by training DNNs based on the task experience of a robot. A task sequence comprising image and motion information can be embedded into a time-series DNN to serve as the dynamics of a sensory-motor sequence. In addition, the task operation can be repeated by acquiring the dynamics in a cyclic form. Further, the process of designing dynamical systems is applied to design a switching system herein.  A dynamical system represents a space of time-series changes in the dynamics of a robot motion based on the environmental information obtained from the camera image. 
In this study, we propose a method to change the transition in the dynamics to deal with motion branching.
\par
To design a dynamical system  with a switching system, it is important to acquire the network dynamics in a switchable form for each subtask. This indicates that the dynamics must comprise a common section in which the internal state of the network is identical  among multiple dynamics since the network determines the subsequent output value depending on the internal state and the given input value. This section is called a "point attractor." Based on the aforementioned framework, \cite{koma}, \cite{kase} created a point attractor that switches between various subtasks by adding constraints to ensure that the internal state remains similar during the initial and final states. In these studies, the robot was made to complete a long task sequence by switching the dynamics of the trained subtask at the point attractor based on the sensory signal of the image from the mounted camera. We use this point attractor to switch between the subtask dynamics.
\par
We extend the work in \cite{kase} to propose a method to design dynamical systems with point attractors that accept (i) instruction signals for instruction-driven switching. To form such point attractors, we incorporate (ii) the instruction phase in task sequences, and it divides the task sequence dynamics into various subtasks.  In this study, only the instruction signals that correspond to motion branching are used. These signals are provided as a condition to determine the next subtask in an ambiguous situation. The instruction phase is a part of the task sequence. We attempt to switch between subtask dynamics using a combination of the sensory and instruction signals at the aforementioned point attractor.  The proposed method uses two DNNs for handling these signals. One DNN of the model autonomously extracts image features from raw images for performing sensory-driven switching, whereas the other DNN designs a dynamical system that self-organizes the relations of signals. To evaluate the switching and generalization ability of the proposed model, we apply it to a cloth-folding task as an example of a soft material manipulation task  that includes motion branching.

\section{Related Works}
To ensure that robots can perform more advanced tasks, it is important to design a framework and task sequences to acquire the required dynamics. This is because combinations of dynamics while learning multiple tasks may be unintended even though  optimizing the dynamics of the target task.
\par
Specifically, deep reinforcement learning is used to perform research on object manipulation owing to its generation precision for a specific task. Robots can learn a trajectory policy and the robot arm actuator torque signals  based on the camera image \cite{DRL1}\cite{DRL2}. These approaches are promising from the viewpoint of performing a specific task with sufficient accuracy. Further, they can be extended for use in multi-task learning if the tasks involved are very similar \cite{DRL3}. However, because the optimized dynamics are limited to the set of learning objectives, the switching system for subtasks is not trained while searching for motion.  Switching systems for separate subtasks are needed only in rare cases.
\par
In some cases, a dynamical system based on recurrent neural network (RNN) is used to perform object manipulation tasks.  The model is built for each task. RNNs use an internal memory to calculate the subsequent output from prior inputs. For robot manipulation, this characteristic of RNN is effective from the viewpoint of processing sequential information and maintaining robustness against some  noise. Moreover, a framework for combining multiple DNNs can be used to integrate multiple types of sensory information  \cite{Noda}. Although such a framework can be used to capture signals for switching motions, the conventional method cannot deal with complicated combinations that involve motion branching.  Motion branching occurs at a point where the internal state and sensory signals are identical in multiple subtask dynamics. At that point, the network cannot determine the dynamics  to shift by sensory-driven switching. For example, in \cite{kase}, a method for automatically switching the subtask only from  visual feedback using the aforementioned framework was proposed. However, it is not possible to handle more complicated motion switching  such as motion branching.
\par
RNNs have been used to acquire specific languages of interactive systems in robotics through sensory-motor learning \cite{Heinrich}\cite{Yamada}; however, it is difficult to apply this method to actual tasks. In \cite{Yamada}, an RNN that could self-organize cyclic attractors reflecting the semantic structure and represent interaction flows using its internal dynamics was used. However, this RNN employed object color centroids as  visual information. Thus, the method cannot handle complicated objects in a real environment, such as soft materials, based on the above information. In addition, the aforementioned method focused only on expressing the semantic relation acquired using a model, and they cannot be used to generate long sequential tasks for a robot with high-dimensional DoF.
\par
The key contribution of our method is that it designs sensory and instruction-driven switching systems for motion branching in dynamical systems formed by DNNs. Further, it enables the execution of flexible object manipulation tasks by training robots with sensory-motor experience comprising camera images and motions of high-dimensional DoF.
\par
We design point attractors for each subtask dynamics. To manipulate the point attractors with feedback from sensory and instruction signals, the model must provide these signals at that point. We divide the task sequences into multiple subtasks and incorporate a section (instruction phase) in which simple vectors (instruction signals) are provided from external sources. Further, we design an instruction phase such that the model depicts almost the same input value at the beginning of this phase. By designing the instruction phase as a point at which the internal state of the network is identical, the dynamics of each subtask is designed as a trajectory attractor with a switchable common section.
\par
To combine two different types of signals, we use a hierarchical RNN that can acquire long and short-term dynamics. Given that instruction signals represent abstract motion instruction in some cases,  the instruction signals represent different subtask motions depending on the situation. Thus, the model must learn task transition from the time-series of sensory signals. If the model can learn the relations of these signals in different internal dynamics, the robot can switch to appropriate subtasks. In  our experiment, we use the "direction of motion" as the instruction signals. We depict that the model acquires appropriate relations of these signals by visualizing the internal dynamics of the network during a task sequence.

\section{METHOD}
We propose a learning-based method consisting of two DNNs to achieve flexible object manipulation with the switching of multiple subtasks during motion branching. The proposed model is depicted in Fig. 1. It is constructed using the following two DNNs: (a) convolutional autoencoder (CAE) for extracting low-dimensional image features  that represent the relationship between the object and robot arm in task manipulation from high-dimensional raw images, and (b) multiple time-scale RNN (MTRNN) to design dynamical systems for sensor- and instruction-driven switching and  to generate the next motion based on  the previous image features and motions.  Our main ideas are as follows:
\begin{itemize}
  \item Design  an "instruction phase" in task sequence to form dynamical system for sensor- and instruction-driven switching using DNNs. 
  \item Switch subtask dynamics based on "instruction signals" that represent the abstract motion instruction.
\end{itemize}
\par
To handle instruction signals, we design the task sequence to consist of an instruction phase such that a robot waits for instruction signals at the beginning of each subtask. 
In this phase, signals other than the instruction signals do not change, their values remain almost constant across all the phases. 
Because the signals provided at the beginning and at the end of each subtask are almost identical, the internal state of the network converges to an almost certain state in the instruction phase. 
At this point attractor, we switch subtasks with certain signals: sensory signals from camera images representing the transition of task sequence and instruction signals explicitly indicating the direction of each subtask motion. 
We trained the MTRNN with a task sequence composed of multiple subtasks for switching from sensory and instruction signals.
MTRNN embeds the instruction signals in the layer representing the fast changing dynamics, whereas the sensory signals in the layer represent the slow changing dynamics. 
\setlength\textfloatsep{5pt}
\setlength\abovecaptionskip{0pt}
\setlength\floatsep{0pt}
\begin{figure}[htpb]
  \centering
  \includegraphics[width=7.9cm]{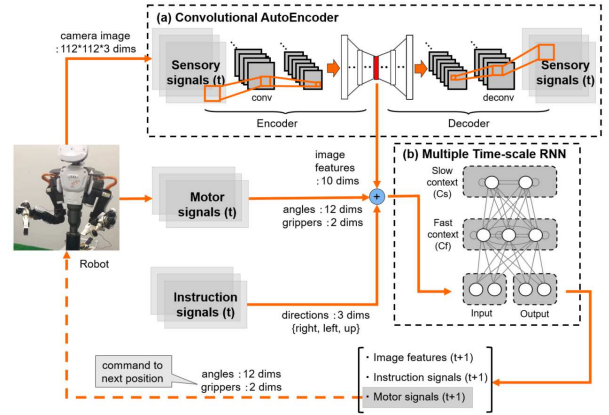}
  \caption{Overview of proposed method with two DNNs.}
\end{figure}
\begin{figure*}[thpb]
  \centering
  \includegraphics[width=15.5cm]{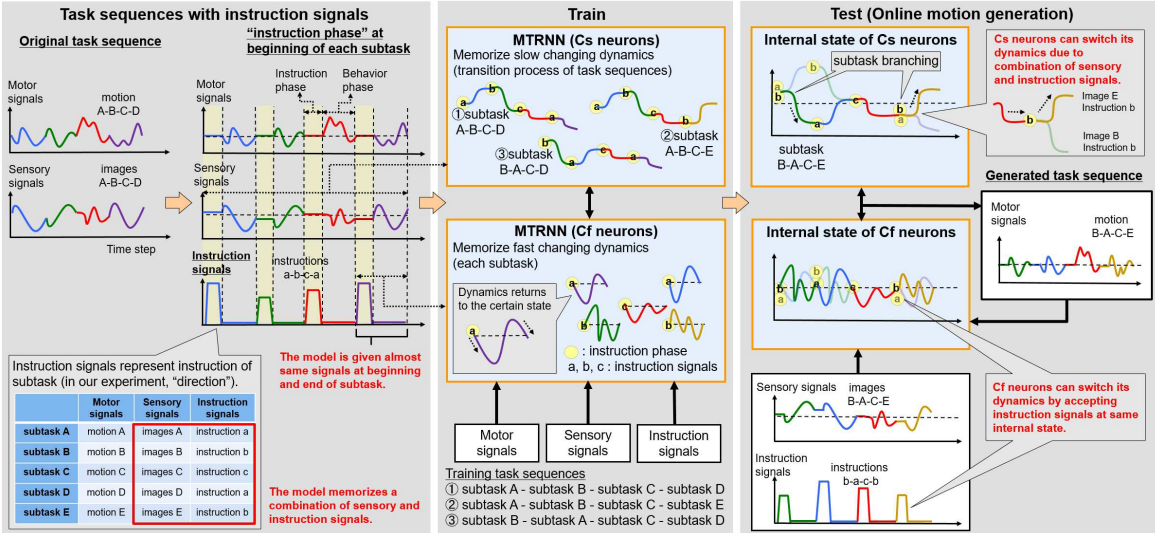}
  \caption{Overview of dynamics acquired by $C_f$ and $C_s$ neurons by designing instruction phase. 
}\end{figure*}

\subsection{Convolutional Autoencoder}
To properly learn the sensory-motor information, appropriate feature extraction from a high-dimensional image is important. It is necessary that the relation between the robot arm and the object being manipulated is reflected in the extracted image features. This is because it affects the generalization performance in terms of generating tasks for unknown object positions or states. Therefore, it is desirable that the feature extractor should be able to handle high-resolution images to the maximum possible extent.
\par
Our model extracted low-dimensional image features from the camera images by using a CAE. 
A CAE comprises a deep autoencoder, convolutional layers, and deconvolutional layers \cite{CNN2}. 
The deep autoencoder, proposed by Hinton et al. \cite{AE}, defines a sandglass-type multilayered fully connected neural network. 
By training the autoencoder to provide output values that are equal to the input values, feature vectors can be extracted at a central hidden layer. 
These encoded feature vectors can be used to represent the state of input data and provide high-dimensional input information using fewer dimensions. 
In our model, we applied convolutional and deconvolutional layers near the input and the output layers, respectively. 
A convolutional layer can handle considerably more input dimensions than that can be handled by a fully connected DNN using fewer parameters \cite{DL1}. 
This enhances the image processing performance by extracting data to different levels of feature maps ranging from edges to partial parts of the image. 
Therefore, the CAE can reconstruct high dimensional inputs into low-dimensional image features.

\subsection{Multiple Time-scale Recurrent Neural Network}
In the proposed model, we implemented MTRNN \cite{MTRNN} to learn the relation between of sensory-motor signals (joint angles, gripper signals, and image features) and instruction signals. MTRNN is a neuro-dynamical model that is used in cognitive robotics as a generation mechanism to predict the subsequent state from the current state. It is composed of three types of neurons: input-output neurons ($IO$), fast context neurons ($C_f$), and slow context neurons ($C_s$). Each type of neuron has a different time constant value. Because of the difference between these values, the dynamics of trained sequences are effectively memorized as combinations of fast changing dynamics in the $C_f$ neurons that have smaller constant values and slow changing dynamics in the $C_s$ neurons that have larger constant values (Fig. 2). 
\par
The propagation of the output of each neuron is limited by the time constants. In forward dynamics, the internal value of each $i$-th neuron at step $t$, $u_i(t)$ is calculated as follows: 
\setlength{\abovedisplayskip}{4pt} 
\setlength{\belowdisplayskip}{4pt}
\begin{eqnarray}
u_i(t)=\left(1-\frac{1}{\tau_i}\right)u_i(t-1)+\frac{1}{\tau_i}\left[\sum_{j \in N}w_{ij}x_j(t-1)\right]
\end{eqnarray}
where $\tau_i$ is the time constant of the $i$-th neuron, $x_j(t)$ is the input value of the $i$-th neuron from the $j$-th neuron, $w_{ij}$ is the weight value of the $i$-th neuron from the $j$-th neuron, and $N$ is the number of neurons connecting to the $i$-th neuron. The respective activation values of the context neurons $c_{i}(t)$ and the output neurons $y_{i}(t)$ are calculated using $sigmoid$ functions. During learning MTRNN, the weight $w$ and the initial value of the slow context neurons $C_s(0)$, are updated using a back propagation through time algorithm \cite{BPTT}. 
\par
From the viewpoint of designing the internal dynamics of the time-series DNN, it is important that each group of neurons learns different dynamics. 
In this paper, the $C_f$ and $C_s$ neurons are assigned to learn different signal information from the time-series sequence. 
In the MTRNN, the $C_f$ neurons obtain more information from the current context, while the $C_s$ neurons from the previous context. 
By using this characteristic, in the proposed method, the $C_f$ neurons represent the dynamics in which it responds to temporary signals such as instruction signals. Moreover, $C_s$ neurons contain information about the transition of task sequences at the point where dynamics branching is involved. 

\subsection{Instruction Phase and Online Motion Generation} 
As mentioned in Section I, to switch the dynamics of the network, each sequence of subtask dynamics must contain a point attractor at which the internal states of the neurons are identical. In addition, the model must be provided with instruction signals at that point. In this study, we attempt to create that point by designing a task sequence that contains sections with regular intervals at which data input is limited. A visualization of our idea is shown in Fig. 2. 
\begin{figure*}[thpb]
  \centering
  \includegraphics[width=15.0cm]{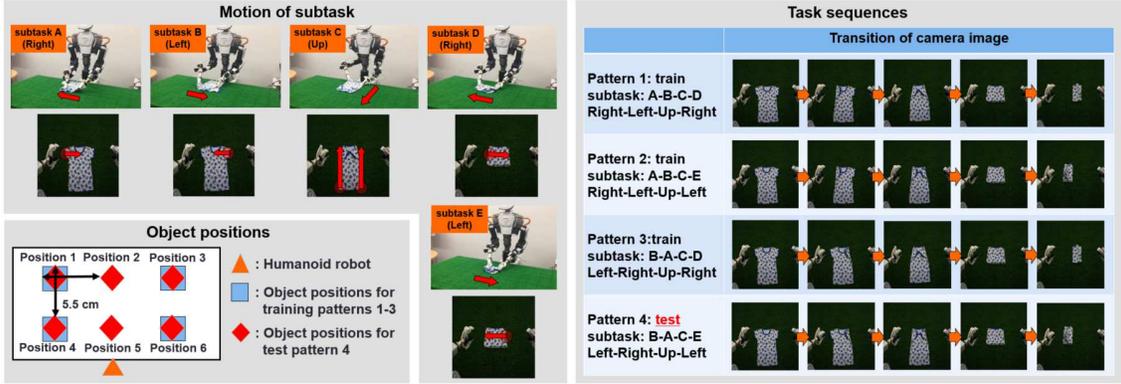}
  \caption{
  Before starting each subtask, the robot receives instruction signals, "Right," "Left," or "Up." 
  The robot manipulates six object positions. 
  We designate folding motions divided from the task sequence as subtask A-E.
}\end{figure*}
\par
In the proposed method, we design the instruction phase such that it divides the original task sequence into subtasks and waits for instruction signal input at the beginning of each subtask. Our model represents the dynamics of each subtask as a trajectory attractor returning to a certain state and switches the subtask to perform the task. We use the trajectory attractor for smooth motion transition. It is difficult to perform smooth and continuous manipulations if the internal state of the MTRNN differs at the beginning of each subtask. This is because the model predicts the next state with strongly being affected from past contexts information.
\par
In the instruction phase, the model is provided with instruction signals that are simple vectors to the MTRNN in the same way as is done in the case of other signals. The input values of the instruction signals are zero in phases other than the instruction phase. The instruction signals are not designed individually for each subtask because they represent the motion "instruction." For example, in Fig. 2, the same instruction signal represents different motions depending on the transition of the task sequence. Therefore, the model must memorize a combination of sensory and instruction signals.
\par
During the instruction phase, the robot maintains a certain position. This is important for restricting the input of motion information and smoothing the transitions between subtasks. After providing the instruction signals, the task sequence transitions to the behavior phase, in which the robot performs subtask. Then, the robot return to a certain position to transition to the instruction phase of the next subtask. 
\par
To form a point attractor, in addition to the constraints of the internal state of MTRNN \cite{kase}, we consider input values at the beginning and the end of each subtask. 
As mentioned above, motor and instruction signals have certain values at this time. 
Because the state of the robot arm in the camera image is constant, camera image in each instruction phase differs only in terms of the shape of the manipulated object. 
And most of the change in the camera image is caused by the manipulation of the robot arm. 
Thus, the model is provided with almost the same input values at the beginning and the end of a subtask. 
Instruction signals are distinguished from other signals, since they are explicitly provided during instruction phase. 
Because $C_f$ neurons memorize fast changing dynamics from the time-series of input data, the internal states of $C_f$ neurons converge to a certain state in the instruction phase and form an attractor point that accept instruction signals. 
Moreover, they are strongly affected by the instruction signals. 
\par
By contrast, $C_s$ neurons expressing the long-term dynamics can learn the transition process of task operation through sensory and motor signals. 
These layers allow the model to learn the relationship between two different signals for switching: sensory signals representing the transitions of multiple subtasks and instruction signals explicitly indicating the instructions of subtask operation. 
By using these signals for dynamic switching, our model can create a switching system without strictly creating instruction signals. 
\par
At the time of task execution, the robot generates motion while acquiring sensory and instruction signals online. When generating a motion, an image from the robot-mounted camera is provided to the CAE, which encodes images into image feature vectors. After combining these vectors with joint angles, gripper signals, and instruction signals, they are provided to the MTRNN. Because of the acquired dynamics that represent the relationship of sensory-motor information, the MTRNN predicts an appropriate output based on a real environment glanced from visual information. The joint angles and the gripper signals predicted by the MTRNN are provided to the robot as commands for the next position. 
MTRNN receives signals as feedback at every time step. Each time step is about 0.15 s. The model can adjust the robot motion in real-time by repeating this process. 
In the instruction phase, the robot is commanded by instruction signals. The $C_f$ neurons can switch their dynamics by accepting immediate input from instruction signals, and $C_s$ neurons can switch their dynamics based on combinations of sensory and instruction signals.

\section{Experiment}
To evaluate whether our model can complete flexible object manipulation tasks by switching subtasks based on sensory and instruction signals, we attempted to complete a garment-folding task with an industrial humanoid robot, Nextage \cite{Nextage}. 
In our experiment, Nextage was commanded to fold a short-sleeved shirt placed in front of it four times. 
\par
Our model was evaluated from the perspective of generalization and interaction ability by using it to make the robot execute a task sequence consisting of untrained subtask combinations and untrained object positions. 
During the trial task, the robot acquired sensory and instruction signals for switching subtask motions. 
By visualizing the internal dynamics of the network, we examined the influence of the instruction phase on a dynamical system. 
\subsection{Design of Task Motion}
The target task consists of the following five subtasks. 
The robot executed these tasks by switching among them. 
This task sequence is represented visually in Fig. 3. 
The robot folded the garment four times. 
The garment-folding task involves motion branching because the place where the robot folds the garment changes owing to the instruction signals. 
The instruction phase was designed at the beginning of each subtask, and the model was provided with instruction signals. 
Each instruction signal corresponded to a folding direction, "Right," "Left," and "Up." 
The task sequence has a few patterns because this instruction phase contains a few possible directions (Table I). 
In the designed task, the second and the third subtasks are uniquely determined, even if there is no instruction signal. 
Because the model is provided with instruction signals in every instruction phase in our experiment, it is expected that the dynamics execute only instruction-driven switching without sensor-driven switching. 
\par
Because the subtask represented by the instruction signal changes depending on the situation, to complete the target task, the robot must appropriately switch among subtasks based on both sensory and instruction signals. 
Instruction signals are not designed individually for each subtask. 
They contain an abstract instruction about folding direction. 
Therefore, in few cases, it is not possible to specify the subtask using instruction signals only. 
For example, in our experiment, the subtask behavior indicated by the instruction signals "Right" of subtask A and subtask D is different. 
Therefore, depending on the task progress, the robot is required to perform sensory- and instruction-driven switching. 
\par

The purpose of this experiment is to verify whether it can switch the subtask at the branching point. 
We prepared training task sequences as minimum required combination. 
The only difference between the test pattern 4 and the training pattern 3 is the last subtask, however, these patterns were predicted as different task since training sequences for MTRNN are series of subtasks. 
It shows that it is possible to extract and combine subtasks without training all patterns of combination of subtasks if test pattern can be executed. 
\begin{table}[hbtp]
  \centering
  \begin{tabular}{c||c|c|c|c}
    \multicolumn{5}{c}{TABLE I: Task Sequence Patterns} \\
    \multicolumn{5}{c}{} \\
    \hline
    Pattern 1 & subtask A & subtask B & subtask C & subtask D \\ 
    (train)   & Right     & Left      & Up        & Right \\ 
    \hline
    Pattern 2 & subtask A & subtask B & subtask C & subtask E \\ 
    (train)   & Right     & Left      & Up        & Left \\ 
    \hline
    Pattern 3 & subtask B & subtask A & subtask C & subtask D \\ 
    (train)   & Left      & Right      & Up        & Right \\ 
    \hline
    Pattern 4 & subtask B & subtask A & subtask C & subtask E \\ 
    (test)    & Left      & Right      & Up        & Left \\ 
    \hline
  \end{tabular}
  \begin{flushleft}
    \ \ \ \ subtask A: fold left sleeve of the garment toward the right direction. \\ 
    \ \ \ \ subtask B: fold right sleeve of the garment toward the left direction. \\ 
    \ \ \ \ subtask C: fold the garment in half from bottom to top direction. \\ 
    \ \ \ \ subtask D: fold the garment in half toward the right direction. \\ 
    \ \ \ \ subtask E: fold the garment in half toward the left direction. \\ 
  \end{flushleft}
\end{table}

\subsection{Experimental Setup}
We performed the experiments to evaluate the results from two viewpoints, namely, "switching ability" and "generalization ability," as described below. 
Interaction ability indicates whether sensory and instruction signals can switch the dynamics of the model appropriately. 
Generalization ability indicates whether our learning-based method can be generalized to any flexible object manipulation task. 

{\bf 1. Interaction Ability:} 
After providing instruction signals, if the model can generate untrained task sequences by switching the subtask of the trained task sequence, it can be said that the model can switch among subtasks according to the instruction signals. 
We trained the model with patterns 1-3, as listed in Table I, as the training task sequences. 
The robot executed pattern 4 as the test task sequence. 
\par
In addition, because three types of instruction signals are used for five types of subtasks in the test sequence (pattern 4,) to combine the subtasks appropriately, it is necessary to integrate sensory and instruction signals. 
In our experiment, we examine whether the hierarchical structure of the MTRNN can acquire the dynamics that represents sensory and instruction signals for the intended motion switching. 
\par
{\bf 2. Generalization Ability:} 
If the model can perform a given task for an untrained object position, it can be said that the model has good generalization ability. 
When we trained the model, it learned four object positions task for each training task sequence. 
The test task sequence was generated by placing the object at six positions, including two positions between the trained object positions, as shown in Fig. 3. 
Hence, we trained our model with 12 patterns (three training patterns $\times$ four object positions).

\subsection{Training and Model Setup}
We trained the proposed model with the training task sequences acquired by operating the robot and using motion capture. After completing the subtask operation, the robot automatically returns to the position it was in at the beginning of the subtask and subsequently starts the operation of the next subtask. Therefore, the motion of each subtask is slightly different but all subtasks have the same initial and final position. 
We smoothed the motions operated by humans as a pre-processing step to effectively train the model. Each subtask in the task sequence comprises 152 steps that consist of an instruction phase (20 steps) and a behavior phase (132 steps). 
The time required to complete the task in approximately 87 s. 
The training pattern described in our paper is simply a pattern of subtasks. 
While training the model, we increased the size of the training set by applying data augmentation and added Gaussian noise and color augmentation to increase the robustness of the CAE and provide a sufficient number of training sets to prevent overfitting. In addition, batch normalization was used in the CAE to improve the learning performance.
\par
We set the parameters of the proposed model to learn the acquired training task sequences. 
The robot has two non-backdrivable six-DoF arms and grippers. 
It captures 112$\times$112 pxs RGB images by using the mounted-camera (total 37,632 dims). 
The CAE extracts 10-dimensional image features from the raw images. 
The values of the instruction signal are [1,0,0], [0,1,0], [0,0,1] for Right, Left, and Up, respectively. 
Therefore, the MTRNN receives 27-dimensional input and output neurons. 
Both the CAE and the MTRNN were trained using mean squared error (MSE) along with the optimizer by Adam \cite{adam}. 
The detailed parameters of the CAE and MTRNN are listed in Table II. 
We searched parameters above by trial-and-error and eventually chose the parameter set that yielded the best results. 
\begin{table}[thbp]
  \centering
  \begin{tabular}{cc}
    \multicolumn{2}{c}{TABLE II: Structure of Networks} \\
    \multicolumn{2}{c}{} \\
    \hline
    Network         & Dims \\
    \hline \hline
    CAE*    & input@3chs - conv@64chs - conv@32chs - \\ 
            & conv@16chs - full@1000 - full@10 -     \\ 
            &  full@1000 - deconv@16chs - deconv@32chs - \\ 
            & deconv@64chs - output@3chs \\ 
    \hline
    MTRNN   & $IO$@27($\tau$:1) - $Cf$@80($\tau$:5) - $Cs$@20($\tau$:70) \\ 
            & $IO$(Joint angles:12, Grippers:2, \\
            &    Image features:10, Instruction signals:3) \\ 
    \hline 
  \end{tabular}
  \begin{center}
  * all conv and deconv filter are stride 2, padding 1 \\
  \end{center}
\end{table}

\section{Results and Discussion}
\subsection{Generation of the Garment-Folding Task}
\begin{figure}[htpb]
  \centering
  \includegraphics[width=6.35cm]{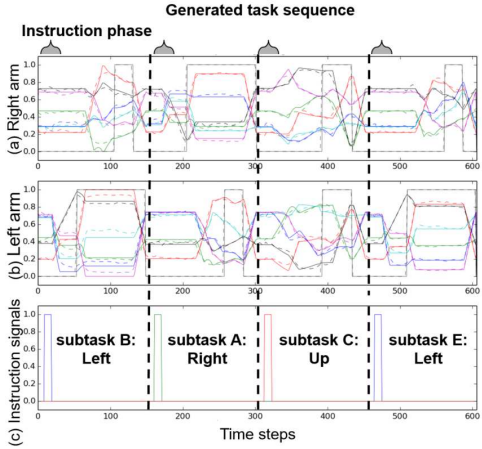}
  \caption{
  Generated test sequences (pattern 4, object position 2). 
  The top part of the figure of the generated test sequence shows motor signals of the right arm (a) and left arm (b). 
  The broken lines indicate the generated output values, and the solid lines indicate the correct values. 
  The bottom part indicates the instruction signals provided to the model (c). 
}\end{figure}
First, we verified the performance of the trained model through online generation. Our model shows some extend of generalization to untrained object positions. We generated untrained sequences for each object position. As an example, the generated untrained pattern sequence (pattern 4) for an untrained object position (position 2) is shown in Fig. 4. In our method, the task performance speed was the same as that of the training sequence because the model needs only forward calculation during online generation. This is promising compared to the model-based method \cite{towel1}\cite{towel2} which tends to be slow when processing high-dimensional data. 
\par
In this experiment, twenty-four trials were conducted with a range of untrained object positions and the robots never failed. This 100\% success rate shows that the robot was able to properly change subtask with instruction signals. 
MSE per joint angle per step averaged over all untrained pattern sequences for untrained object positions was 0.00331. This value corresponds to an 1.63 cm error in the arm-tip position when grabbing the object. 
Although this is the worst result among all generated sequences, it is almost the same as the correct behavior. 
Hence, the model can successfully extract subtasks from the training sequences and combine them. 
\par
In our experiment, even if the robot fails to perform a given task, it tries to continue the task. This is because the training dataset do not include motions to recover from task failures. However, there are some ways to address this problem in our framework. One way is to train motions that recover failures, such as returning clothes to the original position and resuming the folding motion. One of the advantages of our method is that the model can learn multiple motions without specifically designing new behaviors. However, it is difficult to learn all possible failure examples so there is a limit to the error recovery with the above method. Another solution is to repeat a subtask in the case that the robot cannot grip the object because, in this scenario, the state of the object is almost the same as it was before the task was attempted. The internal state of $C_f$ neurons returns to the attractor point and the sensory signals are almost unchanged. Thus, the robot can repeat subtask by providing the instruction signals again.
\par
As mentioned above, the learning-based approach is effective for object manipulation tasks, which are difficult to design. 
However, we need to increase the training dataset for generalization performance over a wider range. 
\begin{figure}[thpb]
  \centering
  \includegraphics[width=5.9cm]{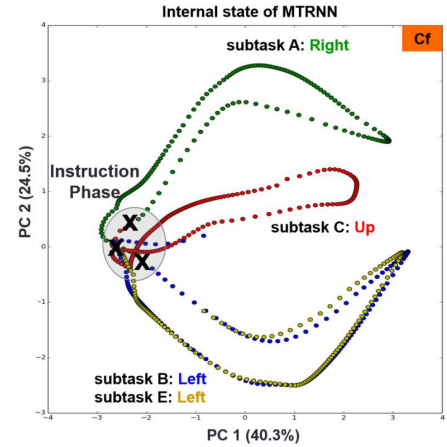}
  \caption{
  Average trajectory of the internal state of $C_f$ neurons. 
  The gray space indicates the instruction phase. 
  The crosses indicate the time at which provision of the instruction signals is initiated. 
}\end{figure}

\subsection{Switching Subtask Dynamics by Instruction Signals}
To confirm the interaction ability of our model, we conducted principal component analysis (PCA) on the internal state of the $C_f$ neurons of the MTRNN (Fig. 5). 
We visualized the average trajectory of the untrained sequences (pattern 4) by projecting them onto the space spanned by the first and second the principal components (PCs). 
Their contribution ratios were 40.3\% and 24.5\%, respectively. 
\par
The dynamics of the subtasks was formed as a trajectory attractor with branch points in the lower layer. 
After the instruction signals were provided, the dynamics of each subtask transitioned to the behavior phase. 
Finally, all dynamics converged on the state that represented the instruction phase. 
This indicates that point attractors consistent with the internal state of the trajectory attractor of each subtask were formed as intended. 
Therefore the model can handle motion branching based on instruction-driven switching. 
\par
Manipulation of the trajectory attractor embedded in the dynamical systems through the sensory-motor experiment with a point attractor is possibly applicable to more complex tasks. 
In this experiment, only one switching phase was designed, but it is possible to design multiple point attractors by adding additional switching phases. 
Moreover, more complex instruction signals such as word vectors can possibly be accepted. 
The results showed that it is possible to explicitly express a complicated task as a combination of simple motion primitives by designing the robot motion as multiple trajectory attractors in dynamical systems. 

\subsection{Integration of Sensory and Instruction Signals}
To confirm whether the proposed model learns the transition of task sequences and to check whether it can combine sensory and instruction signals, we continuously visualized the average value of the internal states of the context layers ($C_f$ and $C_s$ neurons) for each subtask. 
\begin{figure}[thpb]
  \centering
  \includegraphics[width=7.5cm]{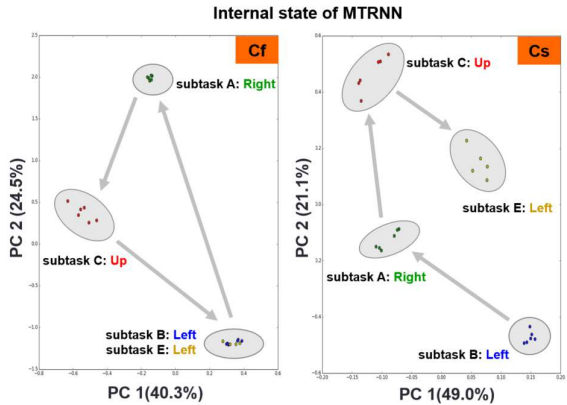}
  \caption{
  Average value of internal states of context layers (PC1-PC2). 
  Each point indicates a subtask B, A, C, and E. 
  The left part of the figure shows the $C_f$ space, and the right part shows $C_s$ the space. 
}\end{figure}
\par
Although the model acquired interaction ability, the $C_f$ neurons did not learn the relationships between sensory and instruction signals. 
The left part of Fig. 6 shows the average value of the internal states of the $C_f$ neurons. 
Each point represents a subtask of an untrained sequence for all object positions. 
These points were projected onto the same space as that shown in Fig. 5. 
Regardless of the object position or shape, three clusters corresponding to the instruction signals can be seen in the space. 
Here subtasks B and E are clustered in the same space despite being related to different motions. 
This means that the information from the instruction signals is embedded in the internal states of the $C_f$ neurons, and consequently, the dynamics respond immediately to the instruction signals. 
However, they could not learn subtle differences between the images of subtasks B and E. 
\par
The $C_s$ neurons played a role in learning the task sequences from the sensory and motor signals, and then our model switched the dynamics with a combination of sensory and instruction signals. 
The center part of Fig. 6 shows the average value of the internal states of the $C_s$ neurons. 
The projected space is spanned by the first and the second PCs with contribution ratios of 49.0\% and 21.1\%, respectively. 
Four clusters corresponding to the sensory and motor signals appear in the space. 
The internal states of the $C_s$ neurons represent the entire transition process of the task sequence. 
This suggests that they learned subtle differences between the images of different subtasks for sensory-driven switching. 
Therefore, the model can adapt to the different situations based on visual feedback and different motions indicated by instruction signals. 
\par
In addition, in subtasks A and C which do not require instruction signals, different clusters are projected. 
Although the model executed instruction-driven switching, the model recognized the camera image. 
Thus, the model can arbitrarily combine sensory- and instruction-driven switching. 
\par
Our method hierarchically self-organized the relationship between sensory and instruction signals within the dynamics of each subtask and exhibited interaction ability. 
And it can acquire dynamical systems that can perform sensory- and instruction-driven switching. 
In this study, it was assumed that the instructor knows how to execute the task and always gives correct instruction signals (i.e., unilateral commands to the robot from the instructor and not considering interactions). To perform more complicated instruction signals, it may be necessary to assume a probabilistic model that allows mutual feedback to change the instruction content. 
\section{Conclusion}
We applied a RNN, which can accept instruction signals to garment-folding task consisting of five subtasks, thereby switching the dynamics of each subtask at the point of motion branching. 
In the proposed method, we designed a trajectory attractor whose instruction phase is the point attractor for acquiring the dynamics of subtasks in a switchable form. 
We verified the generalization ability of the method as well as its interaction ability at the motion branching point by performing tasks with untrained object positions and by visualizing the internal state of the network, respectively. 
The result of the method showed that by applying the proposed method to a robot, we could successfully acquire the relationships between sensory and instruction signals in the hierarchical structure of a RNN and complete a target task by switching the associated subtasks interactively. 
\par
In the future work, we would like to increase the variations and complexity of task sequences, and subsequently increase the variety of instruction signals. 




\begin{thebibliography}{99}
        \bibitem{towel1}J.M.Shepard, M.C.Towner, J.Lei, "Cloth Grasp Point Detection based on Multiple-View Geometric Cues with Application to Robotic Towel Folding," IEEE International Conference on Robotics and Automation, pp.2308-2315, 2010.
        \bibitem{towel2}S.Miller, J.Van Den Berg, M.Fritz, T.Darrel, K.Goldberg, P.Abbeel, "A geometric approach to robotic laundry folding," The International Journal of Robotics Research, vol.31(2), pp.249-267, 2012. 
        \bibitem{SURF}H.Bay, T.Tuytelaars, L.Van Gool, "SURF: Speeded up robust features," Lecture Notes in Computer Science, vol. 3951 LNCS, pp.404-417, 2006.
        \bibitem{DL1}A.Krizhevsky, I.Sutskever, G.E.Hinton, "ImageNet Classification with Deep Convolutional Neural Networks," in Proc. of Advances in Neural Information Processing Systems, pp.1097-1105, 2012.
        \bibitem{DL2}A.Oord, S.Dieleman, H.Zen, K.Simonyan, O.Vinyals, A.Graves, N.Kalchbrenner, A.Senior, K.Kavukcuoglu, "Wavenet: A generative model for raw audio." arXiv preprint arXiv:1609.03499, 2016.
        \bibitem{DL3}D.Silver, A.Huang, C.J.Maddison, A.Guez, L.Sifre, G.van, D.Driessche, J.Schrittwieser, I.Antonoglou, V.Panneershelvam, M.Lanctot, "Mastering the game of go with deep neural netoworks and tree search," Nature, vol.529(7587), pp.484-489, 2016.
        \bibitem{koma}P.C.Yang, K.Sasaki, K.Suzuki, K.Kase, S.Sugano, T.Ogata, "Repeatable Folding Task by Humanoid Robot Worker Using Deep Learning," IEEE Robotics and Automation Letters, vol.2(2), pp.397-403, 2017.
        \bibitem{kase}K.Kase, K.Suzuki, P.C.Yang, H.Mori T.Ogata, "Put-In-Box Task Generated from Multiple Discrete Tasks by a Humanoid Robot Using Deep Learning," Proceedings of 2018 IEEE International Conference on Robots and Automation, 2018.
        \bibitem{DRL1}S.Levine, P.Pastor, A.Krizhevsky, D.Quillen, "Learning Hand-Eye Coordination for Robotic Grasping with Deep Learning and Large-Scale Data Collection," The International Journal of Robotics Research, p.027836491771031, 2016.
        \bibitem{DRL2}S.Levine, C.Finn, T.Darrell, P.Abbeel, "End-to-End Training of Deep Visuomotor Policies," The Journal of Machine Learning Research 17, vol.17(35), pp.1-40, 2016.
        \bibitem{DRL3}K.Bousmalis, A.Irpan, P.Wohlhart, Y.Bai, M.Kelcey, M.Kalakrishnan, L.Downs, J.Ibarz, P.Pastor, K.Konolige, S.Levine, V.Vanhoucke, "Using Simulation and Domain Adaptation to Improve Efficiency of Deep Robotic Grasping," arXiv preprint arXiv:1709.07857, 2017.
        \bibitem{Noda}K.Noda, H.Arie, Y.Suga, T.Ogata, "Multimodal Integration Learning of Robot Behavior using Deep Neural Networks, Robotics and Autonomous Systems," Robotics and Autonomous Systems, vol.62(6), pp.721-736, 2014.
        \bibitem{Heinrich}S.Heinrich S.Wermter, "Interactive Language Understanding with Multiple Timescale Recurrent Neural Networks," 24th International Conference on Artificial Neural Networks, no.8681, pp.193-200, 2014.
        \bibitem{Yamada}T.Yamada, S.Murata, H.Arie, T.Ogata, "Dynamical Integration of Language and Begavior in a Recurrent Neural Netowrk for Human-Robot Interaction," Frontiers in Neurorobotics, vol.10(5), pp.1-17, 2016.
        \bibitem{AE}G.E.Hinton, R.R.Salakhutdinov, "Reducing the dimensionally of data with neural networks," Science, vol.313(5786), pp.504-507, 2006.
        \bibitem{CNN2}M.D.Zeiler, R.Fergus "Visualizing and understanding convolutinal networks," in Proc. of Computer Vision–ECCV, pp.818-833, 2014.
        \bibitem{MTRNN}Y.Yamashita, J.Tani, "Emergence of functional hierarchy in a multiple timescale neural network model: a humanoid robot experiment," PLoS Computational Biology, vol.4(11), pp.e000220-1-e1000220-18, 2008.
        \bibitem{BPTT}P.J.Werbos, "Backpropagation through time: what it does and how to do it," Proceedings of the IEEE, vol.78(10), pp.1550-1560, 1990.
        \bibitem{Nextage}Kawada Robotics: Nextage Open, Available: http://nextage.kawada.jp/
        \bibitem{adam}J.Ba, D.P.Kingma, "Adam: A method for stochastic optimization," in Proc. of International Conference on Learning Representations, 2015.
\end{thebibliography}
\end{document}